# Knowledge Detection by Relevant Question and Image Attributes in Visual Question Answering

Param Ahir[1]*, Dr. Hiteishi Diwanji[2]
*Department of Information Technology*
*L. D. College of Engineering*
*Ahmedabad, India*
[1]ahirparam@gmail.com, [2]hiteishi.diwanji@gmail.com

*Abstract*

*Visual question answering (VQA) is a Multidisciplinary research problem that pursued through practices of natural language processing and computer vision. Visual question answering automatically answers natural language questions according to the content of an image. Some testing questions require external knowledge to derive a solution. Such knowledge-based VQA uses various methods to retrieve features of image and text, and combine them to generate the answer. To generate knowledge-based answers either question dependent or image dependent knowledge retrieval methods are used. If knowledge about all the objects in the image is derived, then not all knowledge is relevant to the question. On other side only question related knowledge may lead to incorrect answers and over trained model that answers question that is irrelevant to image. Our proposed method takes image attributes and question features as input for knowledge derivation module and retrieves only question relevant knowledge about image objects which can provide accurate answers.*

## 1. Introduction

The task of describing visual objects is related to the visual Turing test. Visual question answering is a sub-task of that field. It is a bit complex as it is an AI-complete problem. The ultimate goal of such vision tasks is to be capable to describe images the same as humans.

Visual question answering has a number of use cases. It helps in image retrieval, online chatbots that generate answers from web content, automated driving, object descriptions, food nutrition calculation. VQA has some societal impact as VQA systems are helpful to visually impaired people. They can capture their surrounding in the image and then ask a question about it to get a better understanding of it. Visually impaired farmers can capture the images of their fields and can ask questions like how tall yield has grown? Is any leaf having any disease? Is soil looking less humid? Blind children can ask questions from a storybook. VQA makes possible for a visually impaired person to understand any kind of data visualization.

Visual question answering system can help in humanizing human-computer interactions in the artificial intelligence field in such a way that it becomes similar to human conversations. It is a multi-disciplinary research problem and requires concurrent processing of textual features from a question and visual features from the image It uses NLP to understand the input question and answer it. It is significantly different from the normal NLP problem as it requires analysis and reasoning of text over the content of the image. Object recognition techniques help in identifying the content of the image. To make the process simpler one can derive which areas of an image are important to answer the given question by providing those parts of the question to the image processing module. So that it gives attention to only essential regions of an image and process them only. In VQA system text analysis and image analysis are mutually dependent on each other. As a human, we can easily identify objects, their position and surrounding in an image, understand the question and its relation to image and can use the knowledge and common sense to answer it. When we want a computer system to perform the same tasks systematic approach and algorithms are required. Knowledge aware VQA add one more step to the entire process that is deriving essential knowledge which can help in generating answer. The process of such a VQA system contains four modules, (i) Question features extraction (ii) Image feature extraction (iii) Knowledge extraction (iv)Answering Module. Various deep learning techniques are used to implement these modules. For processing and extraction of text features recurrent neural network (RNN)[1] is used. For processing and extraction of image features convolution neural network (CNN)[2] is used. For knowledge extraction established knowledge bases like concept net[3], DBPedia[4] are used. To predict the correct answer various classification methods are used.

Previous models of knowledge aware visual question answering have some drawbacks. These models either retrieve knowledge about question features or image features which is not always adequate to answer. As if only question-based knowledge is available there is a possibility that generated answer is not accurate for a given picture because image contains attributes that are not inclined with usual world knowledge. For example, if the user asks a question like "What is the color of apple?" then question dependent system will generate answers like red, yellow, green or pink which are common colors for an apple. There is a possibility that the image is abstract and the artist has drawn an apple in blue color then the generated answer will become false. So whenever abstract images are present where everything is upon the imagination of the artist then the system will generate a false answer for question-based system. If only image-based knowledge is derived a then large amount of unnecessary knowledge is extracted which is of no use as the system will derive knowledge about all the objects present in the image regardless of their occurrence in question. Think of an image of market place number of objects present there will be certainly large. Such issues affect the performance of the model and refrain the system from attaining higher accuracy.

Our proposed model uses the concept of attention model in deriving knowledge. This method gives more attention to those image features which are there in the question and derive their knowledge first and after that remaining topmost image attributes are passed to derive knowledge. To derive knowledge, we use the concept of the net knowledgebase. In answer generation model image features, question features and knowledge are passed to generate answers. Using Cross entropy loss function cost is minimized. By implementing and testing the model we derived the correct number of features that need to pass to the knowledge derivation mechanism and how much knowledge should be extracted is derived. Our proposed solution provides better results than all the state of art approaches currently in use.

In summary, the contributions of this paper are, question and image relevant knowledge Detection For VQA, adding question feature matrix to retain image attributes by generating fusion matrix, adding question feature after the attributes prediction layer in such a way that attributes derived at this level gets higher priority than the previously derived attributes. Main Objective is to ensure better classification of Image Attributes.

## 2. Background

Most of the previous work in this field deals with simple visual question answering systems with classifiers. There was no scope to gather peripheral knowledge about objects in question. Previously features of question and image were derived using diverse algorithms. To derive the image features algorithms like VGGNet [5] and GoogleNet [6] were used. To derive textual features algorithms like Word2Vec [7] and one-hot encoding were used. To produce feature fusion, vector representation of the combination of textual and image features were given as input to the multilayer perceptron [8] in neural network. Various methods were used to combine these features like simple concatenation, sum pooling, average pooling or product of features, etc. Answer generation module takes features with the question and feature fusion as input and passes it to classifier to predict the answer. Correct answer is produced by predicting score of candidate answers by relating them to some loss function to measure dissimilarity among two probability distributions. Earlier basic baseline models [9][10] were used to answer the question about the image. Those models answer the question by giving the most frequent answers. Some models even answer the question by randomly picking the answer and then checking its accuracy with various loss functions. Later on, some sophisticated models with a linear classifier or multilayer perceptron were used. Vector representation of the combination of textual and image features are given as input to the multilayer perceptron. Various methods were used to combine these features like simple concatenation, sum pooling, average pooling or product of features, etc. Most of the previous works deal with two models, Simple multilayer perceptron (MLP) and Long short-term memory (LSTM) [11]. MLP used a neural network classifier with two hidden layers. Image features combined with textual features were given as input. To derive the output tanh activation function is used. For textual features representation, a bag-of-words method [13] was used. For image features, the output of the last layer of ResNet [12] (visual geometry group) was used. LSTM model used one-hot encoding for question features and for, image features are derived just like MLP but features are transformed into a linear vector of 1024 dimension to match it with the question feature vector. In both models, for image representation other than ResNet GoogLeNet, VGGNet can be used. The basic problem with using global features is that it generates obscure input space for the model. It is important to attend the most relevant region of the input space. So that generated input space is relevant to the task and model gets clarity about its target area that should be looked upon to generate the answer. An issue with these models is that they include global image features in processing and generation of the answer which is not required. So, the attention model only focuses on local features of the image which are derived using the textual attention model.

## 3. Related Work

Currently, there are several methods in use for external knowledge-based visual question answering. As stated earlier there are four modules and different VQA implement these models differently.

### 3.1. Image Features Extraction

Image features are extracted using various object detection methods to get the list of objects and their attributes. There are many algorithms like, R-CNN [14] which extracts region proposal and pass them into CNN and generate answers using classifier like SVM [15]. There is a Spatial Pyramid Pooling (SPP-NET) [16] which solves the problem of the fixed image size of R-CNN by introducing spatial pyramid pooling layer. There is Faster-RCNN which unlike R-CNN uses regression and classification layer in net and uses the ROI pooling layer [17]. There is a Feature Pyramid Network (FPN) [18] which solves the problem of class imbalance and undetected small objects in faster-RCNN using focal loss and feature pyramid networks. There is RetinaNet [19] which combines the stages of all previous methods into single and generates a single-stage module for object detection. There is YOLO [20] (you only look once) which uses Dark Net architecture to generate answers.

### 3.2. Textual Features Extraction

In VQA text data is questions and answers which are converted into sophisticated feature sets that can be used by classifiers. There are various methods for that like, bag-of-words which counts the presence of words and place it in the feature list. It does not take the number of word occurrences in consideration. There is a TF-IDF [21] algorithm which is related to term frequency and inverse document frequency of words. There is a word2vec algorithm that uses a neural network to generate word embedding in such a way that it preserves the semantics of words.

### 3.3. External Knowledge Extraction

In VQA currently, there is a possibility of extracting knowledge of question or image features from knowledge bases. SPARQL [3] like structured query language is used to generate the query and knowledge is derived.

### 3.4. Answer Generation

This module is divided into two parts (i)feature fusion and (ii)Answer Predictor. For feature fusion currently, the attention model is used. The Attention model contains vectors that assign weights to the regions of the image and words of a sentence. A similar approach is used for the sentence where a specific word can increase the probability of the occurrence of another word. There are many mechanisms available to implement attention like, dot-product, scale dot-product, content-base and location base. Broadly attention mechanism can be divided into three categories, Self-attention – Try to find a relation between input sequences by relating words at a different position in input sequence in each trial. Global-attention – Attend the entire input sequence. Local Attention – Attend parts of the input sequence. Method like Generalized Multimodal Factorized High-Order (MFH) Pooling uses the Multimodal Factorized Bilinear Pooling (MFB) model to provide a fusion of textual and image features. It uses a generalized MFH approach to cascade multiple MFB to find complex correlations to represent an accurate distinction between different question-image pairs. In CROSS-MODAL MULTISTEP FUSION NETWORK (CMF) output from image and word attention is fed as input into CMF network then at each layer three outputs are generated. Out of the two attention features are given for the next CMF unit and fusion feature provides multistep fusion using sum pooling to get the final feature for answer prediction. The process of answer generation is treated as a classification problem. Feature fusion of image and text and derived external knowledge is passed to a multi-layer perceptron. A Softmax activation function is used to generate the probability of each answer. There are some methods that also pass question types in classifiers for better classification of an answer. There are some methods that use pictorial superiority theory and pass image vector externally into the classifier. Some methods that treat the whole problem in terms of a graph and use graph convolution layer to predict the answer. For parameter learning and loss calculation, varied loss functions are used.

### 3.5. Datasets

An external knowledge-based VQA system requires a special kind of dataset that contains questions that requires some external knowledge to generate answers. Normal VQA contains images and question-answer pairs. There are knowledge-based VQA datasets like Ok-VQA, KVQA, and FVQA available.

## 4. Our Approach

Our proposed approach derives question and image feature in the beginning and then passes it into the knowledge extraction module to extract question aware image attributes knowledge which can help in generating highly accurate answers. Our knowledge extraction module is not biased toward textual or visual contents. Extracted knowledge using this method is more precise than older methods.

Our proposed flow is as follows,

1. Region extraction from input image using bounding boxes.
2. Extract image attributes Vatt(I) using Faster R-CNN.
3. Textual data embedding is generated using Glove.
4. Data embedding is passed to LSTM to generate text attributes Tatt (QA)
5. Combine question aware image attributes for knowledge extraction.
6. Write ConceptNet query to derive knowledge about attributes.
7. Vatt(I), Kknow (I, Q) and Tatt (Q) combined into single scene representation as triplet.
8. MLP classifier this triplet as input.
9. Answers with probability distributions are generated.

### 4.1. Image Features Extraction

Images are resized to 224 x 224 dimensions. Such preprocessed images are given to Faster R-CNN pre-trained on ImageNet dataset with VGG16 as the backbone. THE faster R-CNN algorithm detects objects in two parts. In the first part, it detects all the image regions that possibly contain the objects through the sliding window. The first part is called Region Proposed Network (RPN). In second pass object is identified and a vector of dimension 2048 is generated. The feature vector contains the feature and means pooling of the convolution layer or region from where it's detected. Feature files contain the shape of the image and the name of objects.

$$V_k \in d^m \ X \tag{1}$$

$$V_{att}(I_i) = \sum_{k=0}^{n} v^k \tag{2}$$

As shown in equation (1) feature vector of kth objects of single image i contain detected object name m and dimension of its proposed region dm. Such n objects of the same image are combined using their image id. The threshold value for derived objects is between [10,100].

### 4.2. Textual Features Extraction

To create textual data embedding pre-trained Glove word vectors with 42B tokens, 1.9M vocab and 300d vectors are used. Questions and answers are tokenized into words. The vocabulary of the top question and answers are created and word vectors from pre-trained Glove are generated. This vector of dimension 300 is passed to the LSTM network with 1024 hidden layers to generate final output embedding.

### 4.3. External Knowledge Extraction

In this module image attributes Vatt (I) and Question attributes Tatt (Q) are passed into the knowledge extraction module where ConceptNet derives knowledge about objects in JSON REST API. Edges in extracted JSON file contain knowledge about objects in the form of relation and fact pair.

Our approach is to derive the knowledge about those image attributes which are part of question attributes and then trailing the top 5 attributes are considered. Derived knowledge is stored in the form of a triplet and using word2Vector converted in K know (IQ) vectors. Knowledge up to level 11 is derived about an object. Top 11 object edges are selected by the observation that most objects contain at least 11 edges of knowledge available and are enough to generate the answer.

**Algorithm: knowledgeExtraction (I, Q)**

1: for $v^k$ in $V_{att}$ ($I_i$) until count isequal to 5
2:    for m in $v^k$
3:       if (m == $T_{att}$ (Q))
4:          know(m)
5:       else
6:          know(m)
7:          count++
8:    $K_{know}$ ($I_i$, $Q_i$) = know(m) X $I_i$
9:    End loop
10: End loop

In algorithm v^k object vector m is name of object about which know(m) knowledge is extracted about ith image and ith question. It uses concept similar to attention models where textual attention is given to image feature vectors.

**Table 1. Relationship and Knowledge from ConceptNet [3]**

| Relationship | Example |
|---|---|
| RelatedTo | (Skiing,RelatedTo,Mountains) |
| AtLocation | (Tajmahal,AtLocation,Agra) |
| IsA | (Dog,IsA,Mammal) |
| CapableOf | (Dog,CapableOf,barking) |
| UsedFor | (Umbrella,UsedFor,shading) |
| Desires | (Dog,Desires,Playing) |
| HasProperties | (Donuts,HasProperties,sweet) |
| HasA | (Dog,HasA,tail) |
| part of | (Dog,part of, canines) |
| RecievesAction | (Dog,RecievesAction,Fed by human) |
| CreatedBy | (Chocolate,CreatedBy,Coco) |

### 4.4. Answer Generation

Answer generation model is classification model with multilayer perceptron where top N-most answers are considered N-classes. Output is in form of probabilities of each class and for that SoftMax activation function is used. Input to neural network is image feature vector of dimension 2048 is concatenated with question feature vector of dimension 300and knowledge vector $K_{know}$ (IQ). In model training to update the model weights iteratively AMSGrad variant of adam optimizer with 0.003 learning rate is used. It uses categorical cross entropy loss function to calculate to calculate the cost of network.

Z' = concate ($V_{att}$(I) U $K_{know}$ (I, Q) U $T_{att}$ (Q))  (3)

Y' = SoftMax (Z' + $W_{i, q}$)  (4)

Loss calculation,

Loss (Y, Y') = $\sum_{i=0}^{n}(Y_i * \log(Y_i'))$  (5)

Here $W_{i, q}$ is weights for the model and Y' is predicted output and Y is actual output.

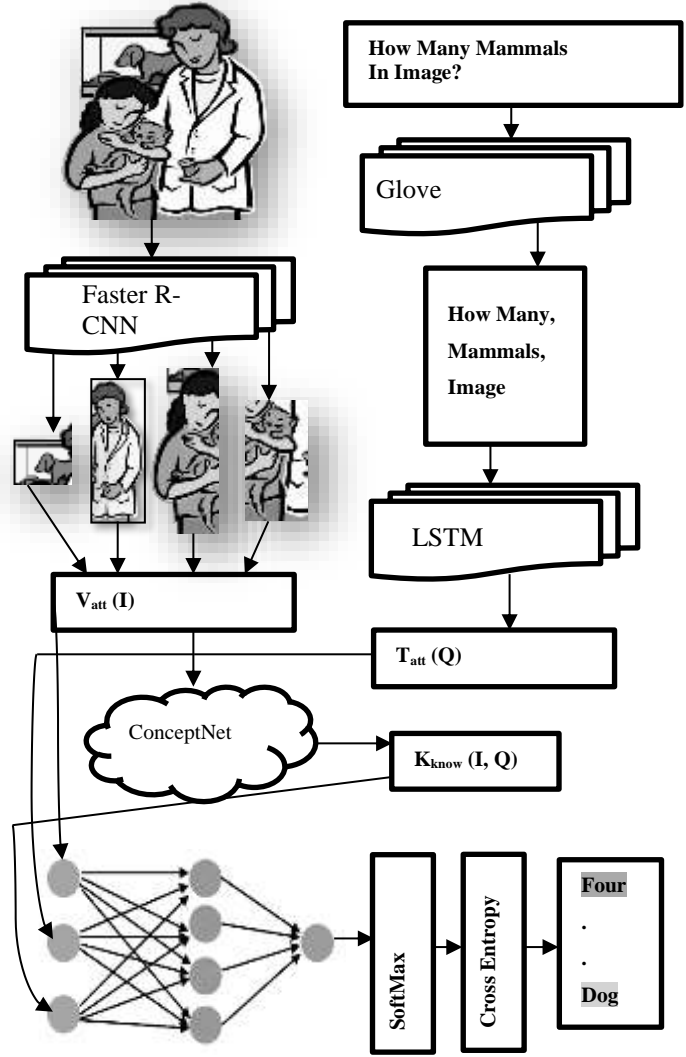

**Figure 1. Proposed Model**

## 5. Experiment

In this section, we evaluate the performance of our model with the state-of-art model. We are using Tensorflow and Keras to implement our proposed solution on the established dataset and compare it with previous results.

### 5.1. Datasets

For training we used Ok-VQA [22] dataset and compared results with model with only image-based model. We also created a small dataset of 20 images and 30 question-answer with knowledge base file to decide different parameters and accuracy of our model. Ok-VQA Dataset contains around 14,055 open-ended questions and 5 ground truth answer per question on an average. All questions in the dataset required external knowledge to retrieve answer.

### 5.2. Evaluation

Our model is trained on 9009 sample and for validation 7779 samples were taken. To evaluate model Wu-Palmer Similarity (WUPS) [23] accuracy metric is used which can calculate the similarity index between two words and return relatedness score between them.

The first step is to preprocess textual and visual data. In textual preprocessing data is converted into lowercase. Question and answers are tokenized into words of maximum character length 20. For image data preprocessing all images are converted into 'bgr' format to make it compatible with OpenCV. All images are resized to [224,224] shape. The second step is feature extraction from textual and visual content. For text feature extraction questions and answers are encoded. In both datasets, the question is stored in the JSON file. Format of this JSON file is as follows, e.g.,

{

"image_id": 81721,

"Question": "How old do you have to be in Canada to do this?"

"question_id": 817215

}

Answers are stored in an annotation JSON file where two additional nodes answer and answer_id are added. After preprocessing this data is converted into tokens and vocab of top words is generated. In our dataset Vocab contains around 45 words. In our dataset top, 15 answers are taken and in the ok-VQA dataset top, 2000 answers are taken.

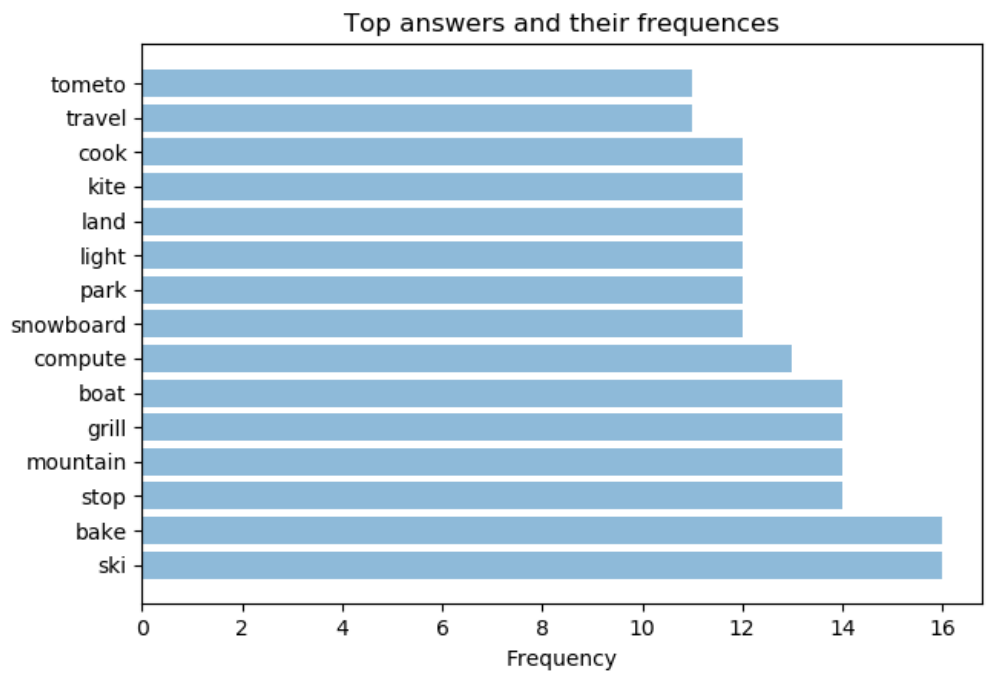

**Figure 2. Top answers in OK-VQA**

After generating vocabulary using glove pre-trained model questions and answers are encoded. For image feature extraction faster R-CNN is used. The objects JSON file is generated. In the knowledge extraction module embedded question-image features are passed. The first common question-image features are passed and then 12 more objects from the image feature file are passed in the knowledge extraction module. ConceptNet python API is used to extract knowledge. Knowledge up to the depth of 11 is extracted. To decide the number of objects to pass and depth of knowledge our dataset is used with a different value. Generated results tell that using 12 objects can help in improving model performance and until depth 11 intelligent and useful knowledge is derived. Format of the derived JSON file is as follows, e.g., {   know_id: 81721,

uri: ConceptNet/e/655e2da1f472ca894742d4156a8d363b

Labels: umbrella, sunny

Surface: "[Umbrella] is used for shading in [sunny] place."

Relation: "usedfor"}

Finally, simple multilayer perceptron and long short-term memory are used for answer classification. In MLP number of hidden units are 1024 and the number of hidden layers is 3. We are using a dropout of 0.5. The activation function for the first and second layers is tanh and Relu. In LSTM number of hidden units are 512 and the number of hidden layers is one. For the final layer of MLP SoftMax activation function with 1000 classes is used to generate answers with their probability. It uses an adam optimizer with an o.oo1 learning rate. For loss calculation, a categorical cross-entropy function is used. We trained our model for 10 epochs with batch size 100. Comparisons of our result with the standard method are given below,

**Table 1. Overall accuracies on training and validation samples of Ok-VQA**

| Model | Accuracy | |
|---|---|---|
| | *Training* | *Validation* |
| Img-attention | 0.6532 | 0.5162 |
| Que-attention | 0.6023 | 0.5170 |
| **Img-Que Co-attention (Proposed)** | **0.6698** | **0.5963** |
| Img-Que Stacked Co-attention [24] | 0.6703 | 0.5972 |

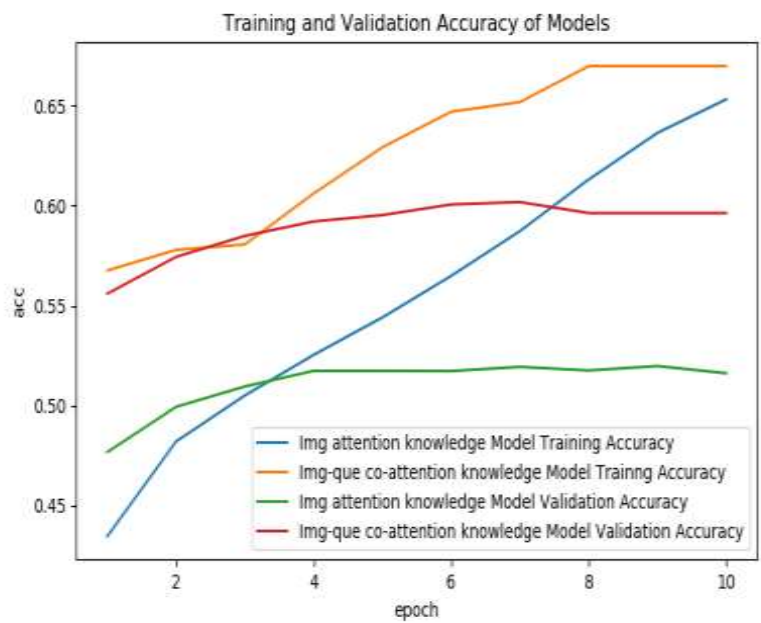

**Figure 3. Training and Validation Accuracy on OK-VQA**

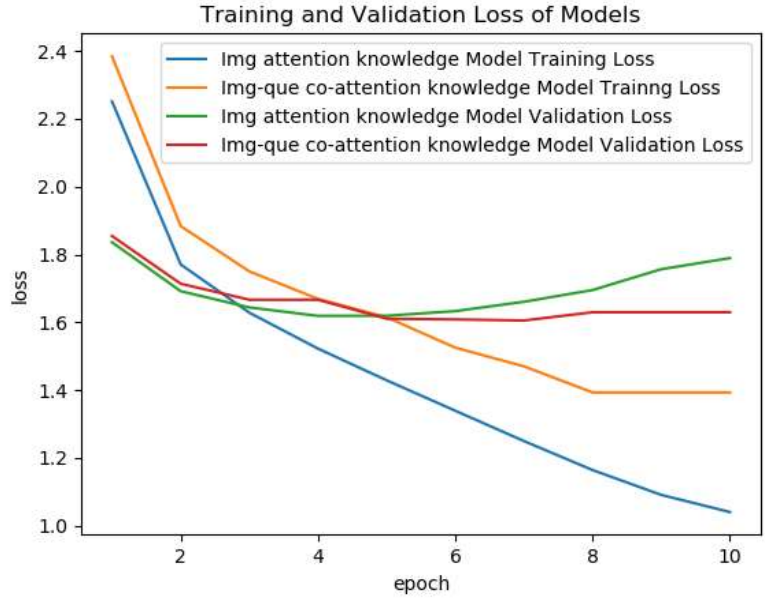

**Figure 4. Training and Validation Loss on OK-VQA**

| | | | |
|---|---|---|---|
| Question 1- How many mammals in the image? | Question 2- How this taste? | Question 3- What is the use of object on table? | Question4- Which flower is this? |
| Answer – 4<br>GT - 4 | Answer – Sweet<br>GT - Sweet | Answer – cutting<br>GT - cutting | Answer – Daffodil<br>GT - Daffodil |

**Figure 5. Examples of Generated Correct Answers**

Figure. 5 shows that accuracy is not much improved on training data because we are still considering dataset ground-truth answers as the first choice of answer. If the dataset does not contain an answer then the only knowledge-based answer is considered. Overall model accuracy is mentioned in table II. Our model accuracy is around 66% for training data and 59% for validation data. We also tried to implement our model on the stacked co-attention VQA model to improve accuracy only marginally. Figures. 6 and 7 show some of the results derived from our model. Results in figure 7 have a probability of less than 50% which is not considered a proper answer. Reason for low accuracy in question 3 is complex grammatically structure of question and in question 4 is knowledgebase is not able to derive knowledge accurately.

## 6. Conclusion

Visual Question answering is the task of forming strong AI for the purpose of building a Visual Turing test for computer vision systems. VQA uses various methods of computer vision and language processing for understanding visual and linguistic inputs and deducting their implication to deliver fitting answers. Some previous systems required additional knowledge that is fulfilled by adding knowledge from the knowledge bases in the system. Previous systems fed visual attributes to the knowledge detection module to derive the information regarding an object in question. The problem with this approach is, it contains lots of additional information about various objects of an image that is not required for answer generation and does not contain the information required for answers. The previous model provides low accuracy for a question like 'why' which heavily depends on the knowledge. To solve this issue our model fed those visual attributes of input in the knowledge detection module that are present in question or at least give them precedence over other visual features. Our model will generate more relevant knowledge to answer questions which eventually increase the accuracy of the model. The previous model provides an accuracy of around 52% which we improved to 59%.

There are many challenges in a knowledge-based approach that gives scope for future work like, question embedding of a model can be more semantically precise for questions like question 3 of figure 7. Knowledge extraction is still not completely automated and can be trained separately to be used in different areas. Domain-specific datasets can help in creating domain-based VQA systems. The process of knowledge classification for answer generation can also be improved. These are some of the problems of the current approach which gives direction for future work in this field.

## Acknowledgments

We like to thank the team of Ok-VQA for generously making their dataset available for use for researchers like us.